\documentclass[runningheads]{llncs}

\usepackage{eccv}

\usepackage{eccvabbrv}

\usepackage{graphicx}
\usepackage{booktabs}

\usepackage[accsupp]{axessibility}  

\usepackage[pagebackref,breaklinks,colorlinks,citecolor=eccvblue]{hyperref}

\usepackage{orcidlink}

\usepackage{bm}
\usepackage{pifont}
\usepackage{multirow}
\usepackage{algorithm}
\usepackage{algpseudocode}

\newcommand{\hsp}[1]{@{\hspace{#1\tabcolsep}}}

\newcommand{\cmark}{\ding{51}}
\newcommand{\xmark}{\ding{55}}
\def\ours{MOD-UV}
\def\mm{\textit{Moving2Mobile}}
\def\ls{\textit{Large2Small}}
\def\ARL{AR$_{\text{L}}$}
\def\ARM{AR$_{\text{M}}$}
\def\ARS{AR$_{\text{S}}$}
\def\APL{AP$_{\text{L}}$}
\def\APM{AP$_{\text{M}}$}
\def\APS{AP$_{\text{S}}$}
\def\gray{\color{gray}}

\begin{document}

\title{\ours{}: Learning Mobile Object Detectors from Unlabeled Videos}

\titlerunning{\ours{}}

\author{Yihong Sun \and
Bharath Hariharan}

\authorrunning{Sun and Hariharan}

\institute{Cornell University}

\maketitle

\begin{abstract}
Embodied agents must detect and localize objects of interest, \eg traffic participants for self-driving cars.
Supervision in the form of bounding boxes for this task is extremely expensive.
As such, prior work has looked at unsupervised instance detection and segmentation, but in the absence of annotated boxes, it is unclear how pixels must be grouped into objects and which objects are of interest.
This results in over-/under-segmentation and irrelevant objects.
Inspired by human visual system and practical applications, we posit that the key missing cue for unsupervised detection is \emph{motion}: 
objects of interest are typically \emph{mobile objects} that frequently move and their motions can specify separate instances.
In this paper, we propose \ours{}, a \textbf{M}obile \textbf{O}bject \textbf{D}etector learned from \textbf{U}nlabeled \textbf{V}ideos only.
We begin with instance pseudo-labels derived from motion segmentation, but introduce a novel training paradigm to progressively discover small objects and static-but-mobile objects that are missed by motion segmentation.
As a result, though only learned from unlabeled videos, \ours{} can detect and segment mobile objects from a single static image. 
Empirically, we achieve state-of-the-art performance in unsupervised mobile object detection on Waymo Open, nuScenes, and KITTI Datasets without using any external data or supervised models. 
Code is available at \href{https://github.com/YihongSun/MOD-UV}{\texttt{github.com/YihongSun/MOD-UV}}.

\end{abstract}

\begin{figure}[hb!]
  \centering
  \includegraphics[width=0.926\textwidth]{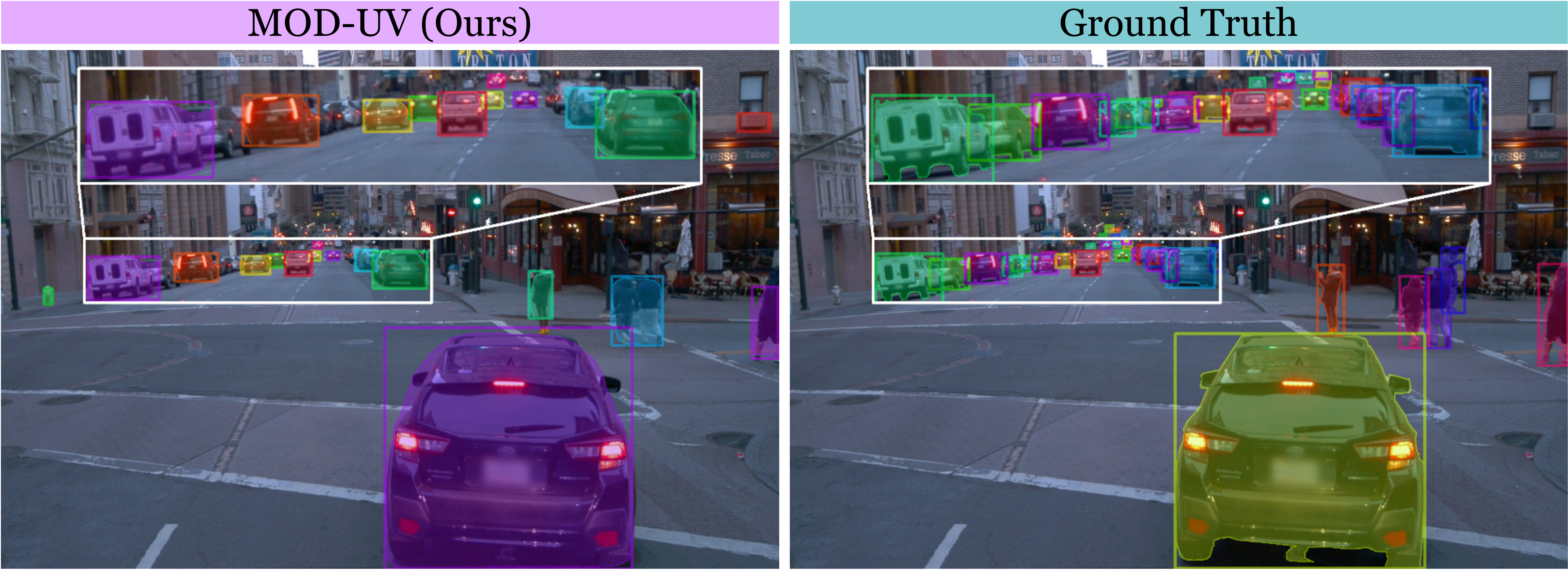}
  \caption{Our approach, \ours{}, learns from unlabeled videos in Waymo Open~\cite{waymo} only and can reliably detect and segment mobile objects from a single input image.}
  \label{fig:teaser}
\end{figure}
\section{Introduction}
\label{sec:intro}
Embodied agents such as self-driving cars must detect and localize objects of interest such as traffic participants to operate safely and effectively.
Today, building such a detector requires the expensive and laborious annotation of millions of boxes over thousands of images.
This process is so expensive that the largest detection dataset is orders of magnitude smaller than classification datasets and has much fewer classes.
The limited set of classes further runs the risk of missing important object categories (\eg snowplows for self-driving applications).

These concerns have motivated research into \emph{unsupervised} object detection techniques that automatically discover objects from unlabeled data~\cite{wang2022freesolo, cutler, hassod}.
Under the hood, these techniques use self-supervised features to segment unlabeled images and produce candidate object annotations which are then used to train a detector.
However, while promising, these approaches often produce many uninteresting and irrelevant ``objects'' in cluttered scenes (\eg buildings and roads) and over- or under-segment objects of interest (\eg multiple detections partitioning a large bus or a single detection grouping a row of parked cars).
These failures shouldn't be surprising: after all, how can a completely unsupervised feature representation encode which assortment of windows, doors and wheels belongs together as an object, and which objects are of interest?

In this paper, we argue that a key missing cue for addressing the aforementioned issues in unsupervised instance detection is \emph{motion}.
In a practical sense, objects of interest are commonly \emph{mobile} objects that frequently move. 
For example, robots performing navigation tasks must plan their trajectories carefully around objects that \emph{can move} of their own volition.
Thus, we argue that if we see similar groups of pixels frequently move of their own volition in unlabeled videos (\eg vehicles and pedestrians in driving videos), this is sufficient information for building a \emph{mobile object detector} that can detect such instances in static frames.

The importance of motion as a perceptual cue (the Gestalt principle of \emph{common fate}) is well known~\cite{palmer2003}.
Indeed, motion-based grouping is one of the first forms of grouping to appear developmentally in human infants~\cite{spelke1990principles} and can bootstrap other grouping cues~\cite{ostrovsky2009visual}.
There is also some prior work on using motion-based grouping in computer vision to produce (pseudo) ground-truth for feature learning~\cite{pathak2017learning} and discovering isolated, salient objects~\cite{croitoru2017unsupervised, choudhury2022guess}.
However, there is a big-gap between motion segmentation and the kind of ground-truth we need for training a full-fledged instance-level object detector.
First, motion segmentation produces a binary segmentation; this must be resolved into individual instances.
Second, it only identifies moving objects and does not include objects (\eg parked cars) that are static but mobile.
Finally, motion segmentation only identifies nearby objects, since the pixel motion of faraway objects is too subtle to discern.
Thus motion segmentation alone will still under-segment and miss many mobile objects: a problem for building object detectors.


Here we propose a new training scheme to address these challenges.
Our approach (\ours{}, a \textbf{M}obile \textbf{O}bject \textbf{D}etector learned from \textbf{U}nlabeled \textbf{V}ideos only; \Cref{fig:teaser})
 trains on unlabeled videos alone and produces a \emph{mobile object detector} that can run on static frames.
We first generate pseudo training labels from motion segmentation estimated by our prior unsupervised framework Dynamo-Depth~\cite{sun2024dynamo}.
We then propose a new training scheme to address the challenges above, resulting in a final mobile object detector that detects $12\times$ more mobile objects than the initial motion segmentation.

We test \ours{} on self-driving scenes but evaluate on a variety of datasets.
Specifically, we compare to recent state-of-the-art unsupervised object detectors and demonstrate improvements across the board, with notable improvements in Box AR by 6.6 on Waymo Open \cite{waymo}, 4.9 on nuScenes \cite{nuscenes} and 6.2 on KITTI\cite{kitti}.

In sum, our contributions are:
\begin{enumerate}
\item We argue that motion as a cue is sufficient for unsupervised training of instance-level object detectors. 
\item We propose a new training scheme that trains on unlabeled videos to produce a mobile object detector that can run on static images.
\item We demonstrate marked improvements over unsupervised object detection baselines across a range of datasets and metrics.
\end{enumerate}

\section{Related Work}
\label{sec:related}

\subsubsection{Unsupervised Object Detection/Discovery from Images.}
Learning to identify and localize objects from unlabeled images is a challenging task, since object information must be obtained without any explicit human annotations. 
A long line of work seeks to discover prominent objects in large image collections~\cite{cho2015unsupervised, vo2019unsupervised, vo2020toward, vo2021large}.
However, these approaches are fundamentally limited by the quality of the object proposals.
More recently, Locatello \etal \cite{locatello2020object} and DINOSAUR \cite{dinosaur} consider object discovery as object-centric learning \cite{greff2020binding} and decompose a complex scene into independent objects. 
Nevertheless, the reconstruction objective is difficult to scale and likely to discover irrelevant patches as well.

More recent work has relied on the fact that bottom-up segmentation algorithms when applied to self-supervised pretrained representations yield good object proposals.
Specifically, pseudo mask labels can be generated from DINO \cite{dino} features to train downstream object detectors~\cite{simeoni2021lost, simeoni2023found, wang2022freesolo}.
MaskDistill~\cite{van2022discovering} extends upon this by distilling from affinity graph produced by DINO~\cite{dino} features, while TokenCut \cite{wang2023tokencut} and CutLER \cite{cutler} use Normalized Cuts \cite{shi2000normalized}.
HASSOD \cite{hassod} leverages hierarchical adaptive clustering, which improves the detection of small objects and object parts. 
This line of work now produces detectors that can run on static images, similar to our work.
However, the detected objects can often be irrelevant (\eg buildings and road) or over-/under-segment objects of interest.
In contrast, our proposed \ours{} discovers and detects a more meaningful and practical set of mobile objects instead, and can learn from their apparent motion in unlabeled videos only without relying on any additional datasets.

\subsubsection{Unsupervised Object Detection/Discovery from 3D.}
In addition to unlabeled images, 3D information is also useful for discovering objects.
Herbst \etal \cite{herbst2011rgb} and MODEST \cite{modest} discover non-persistent objects via multiple traversals with 3D sensors.
Garcia \etal \cite{garcia2015saliency} discovers salient objects by late-fusing color and depth segmentation from RGB-D inputs, while Tian \etal \cite{tian2021unsupervised} generates candidate segments from LiDAR 3D point clouds.
In comparison, \ours{} does not require any additional sensors or modalities beyond unlabeled videos, which allows our method to work in more general settings.

\subsubsection{Unsupervised Object Detection/Discovery from Videos.}
Inspired by Gestalt principle of common fate~\cite{palmer2003}, another class of related work discovers objects via their apparent motions observed in videos~\cite{yan2019semi, yao2020video, lu2020learning}.
By leveraging optical flow information from an input video, a binary segmentation of the moving objects can be extracted~\cite{pathak2017learning, kwak2015unsupervised, xie2019object, lamdouar2020betrayed, yang2021self, zhang2021deep, zhou2021target, singh2023fodvid}. 
Lian \etal \cite{lian2023bootstrapping} proposes further improvements for cases of articulated/deformable objects and shadow/reflections by relaxing the common fate assumption.
Another line of work uses a reconstruction objective to identify the moving object~\cite{tangemann2021unsupervised, aydemir2024self}. 
Du \etal \cite{du2020unsupervised} models explicit object geometry and physical dynamics by exploiting motion cues.
Bao \etal \cite{bao2022discovering} improves training for object-centric representation via an additional motion segmentation regularization, while SAVi++ \cite{elsayed2022savi++} incorporates LiDAR data when training an object-centric video model.
Unlike \ours{}, these approaches do not build a static image detector. However, the output segmentation can be used as an initialization for our approach.

Closer to our work, Pathak \etal \cite{pathak2017learning}, Croitoru \etal \cite{croitoru2017unsupervised} and Choudhury \etal \cite{choudhury2022guess} train a single-frame binary segmentation network on video frames as input and leverage object motion as supervision. Furthermore, LOCATE \cite{singh2023locate} applies graph-cut to obtain binary motion mask from DINO \cite{dino} and optical flow feature similarities, which in turn is treated as pseudo-labels for bootstrapped self-training of a downstream segmentation network. 
However, these techniques can only detect a \emph{single} salient object per frame.
In contrast, \ours{} generalizes to multi-object detection beyond single-object saliency detection.

\newcommand{\Lzero}{L^{(0)}_i}
\newcommand{\Lone}{L^{(1)}_i}
\newcommand{\Ltwo}{L^{(2)}_i}  
\newcommand{\dL}{\bm{L}} 
\newcommand{\dS}{\bm{S}}  
\def\HG#1{\overrightarrow{#1}}

\section{Method}
\label{sec:method}

\subsubsection{Problem setup:} 
We assume an uncurated collection of unlabeled videos as input.
In particular, we assume that these videos are obtained by an embodied agent observing, and optionally acting in the world. 
Solely from the unlabeled videos, the goal is to learn a detector that operates from a single frame and can detect and segment all mobile objects that can move of their own volition.

\subsection{Initialization with Unsupervised Motion Segmentation} 
\label{sec:method-init_with_mask}
A key insight in \ours{} is that if an object \emph{can move}, it is likely that it \emph{does move} many times in the collected data.
Thus, we start by identifying moving objects in the videos; they can be initial seeds for learning about mobile objects.

Fortunately, the task of identifying independently moving pixels from unlabeled videos is a well-studied one\cite{bouthemy1993motion, shi1998motion, yan2006general, ranjan2019competitive}.
In particular, many recent techniques have been proposed that learn motion segmentation without supervision from unlabeled videos.
Many of these techniques also produce depth and camera motion \cite{ranjan2019competitive,luo2019every,li2021unsupervised,hui2022rm}.
Here, we use our prior work, Dynamo-Depth~\cite{sun2024dynamo}.
Dynamo-Depth trains on unlabeled videos and learns both a monocular depth estimator as well as a motion segmentation network.
We use the outputs of these trained networks on our unlabeled videos as a starting point.
Concretely, we denote the input set of unlabeled videos as $\{v_i\}$, with each video $v_i$ containing consecutive frames $I_1, \dots, I_n$ and known camera intrinsics. 
For each frame $I_i$, we obtain its estimated motion mask $m_i$ and estimated monocular depth $d_i$.

With the given binary motion mask $m_i$, we first need to partition the moving pixels into instance-level labels.
While disjoint moving regions can be easily separated, multiple moving objects in the same region would require additional information (\eg 3D information) to separate.
Therefore, for each image $I_i$, we project the corresponding moving pixels in $m_i$ into pseudo 3D point clouds $P_i$ via the estimated monocular depth $d_i$ and inverse camera intrinsics $K^{-1}$.\footnote{$\HG{\cdot}$ denotes the conversion to homogeneous coordinates}
\begin{equation}
    P_i = \{ d_i(p) K^{-1} \HG{p} \quad | \quad m_i(p) = 1 \}
\end{equation}
Then, we cluster $P_i$ via DBSCAN \cite{dbscan} to get a pseudo depth-aware partition of the motion mask $m_i$, which we treat as the initial pseudo-labels, $\Lzero$.

\begin{figure}[tb]
  \centering
  \includegraphics[width=\textwidth]{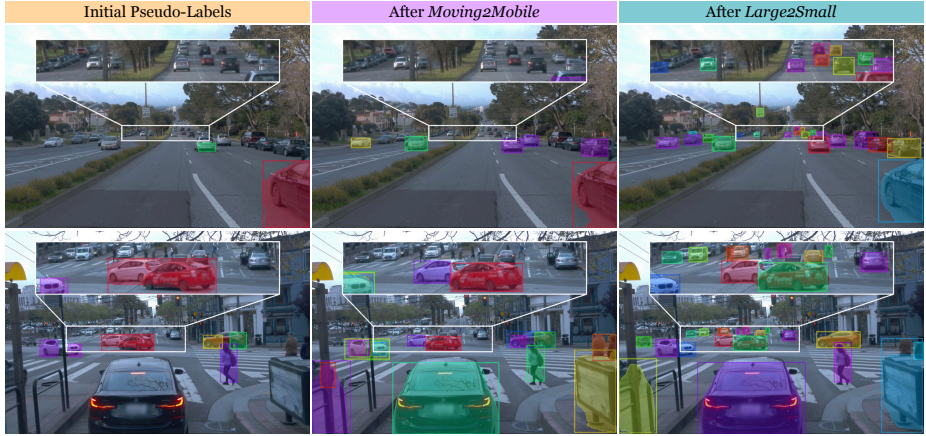}
  \caption{Visualization of pseudo-labels at each stage of our self-training paradigm. From the initial pseudo-labels $\Lzero$ generated from motion mask, $\Lone$ retrieves the large static objects after \mm{} and $\Ltwo$ recovers the small objects after \ls{}.}
  \label{fig:pipeline}
\end{figure}

We evaluate the quality of these pseudo-labels qualitatively in 
\Cref{fig:pipeline} and quantitatively in the top rows of \Cref{tab:ablate_pseudo_mask}.
We find that these pseudo-labels have high precision, but have two severe limitations.
First, they only identify moving objects, so they miss objects that are static but can move (\eg parked cars)
Second, they miss almost all faraway objects which tend to be small.
This is because the apparent pixel motion of faraway objects is very hard to detect.

To tackle this issue of limited recall, we propose two self-training stages, \mm{} and \ls{}, that progressively recover more mobile objects in the scene by aligning the training distribution of \emph{static} and \emph{small} objects with the available large moving objects in the initial pseudo-labels, respectively.

\subsection{Self-Training for Unsupervised Mobile Object Detection.}
\label{sec:method-self_training}
Here, we describe each self-training stage of \ours{}, as we progressively discover mobile objects to train the final mobile object detector.

\subsubsection{\mm{}: Learning to Detect Static Objects.}  
On the left of \Cref{fig:pipeline}, the initial pseudo-labels $\Lzero$, while having high precision, fail to capture the large static objects, \eg the black sedan in the bottom. 
However, a parked black sedan looks the same as a moving black sedan if all one has is a single frame.
In other words, moving and static objects are indistinguishable when observed from a single frame.

Thus, in \mm{}, we simply train a detector to reproduce the pseudo-labeled instances in  $\Lzero$, but with only a single frame as input.
Since object motion is not apparent in a single frame, this detector cannot distinguish moving objects from static ones and thus is forced to detect anything that share the appearance of moving objects, thus detecting static mobile objects as well.

However, there may exist domain-specific statistical regularities that give hints to object motion even in a single frame.
For example, lit-up tail-lights might indicate that the car is stopped, while a highway background might suggest that the cars are moving.
To prevent the detector from overfitting to these priors, we stop training early.
Afterwards, we treat the high confidence predictions by the detector as the pseudo-labels for the next round $\Lone$.

\subsubsection{\ls{}: Learning to Detect Small Objects.}  
As shown in \Cref{fig:pipeline}, $\Lone$, pseudo-labels after \mm{}, appropriately recovers the large static objects, however, the smaller objects remain absent.
Intuitively, faraway objects have much smaller apparent pixel motion (and thus are absent from $\Lzero$) and also look different from large moving objects (and thus are absent from $\Lone$).

To learn to detect small objects, we create a new training dataset by scaling down both the image and the pseudo-labels (while also padding the image to maintain image size).
We then train a separate ``small object'' detector by training on this new dataset.
Intuitively, by training on the scaled down training pair, the output detector would need to detect the same object at a much lower scale, directly promoting the extension to small objects.

Also, since $\Lone$ came from a heavily-regularized detector, we maintain and finetune the pseudo-labels for larger objects by training a second detector from scratch in parallel, on the training pair $(I_i, \Lone)$ without down-scaling or padding.
Notably, this is different from traditional scale jittering, since the singular detector would be discouraged from detecting small objects at larger scales due to the limitations in $\Lone$.
Upon convergence, we have a Large-object detector trained at original scale and another Small-object detector trained at a reduced scale. After aggregating their predictions and resolving conflicting proposals, we obtain the final pseudo-labels, $\Ltwo$.
We note that separating out large and small object detectors in this way has been explored in supervised face detection~\cite{tinyfaces}.

\subsubsection{\textit{Final Round} of Self-Training.} 
As shown in the right of \Cref{fig:pipeline}, $\Ltwo$, the final pseudo-labels after \ls{}, successfully recovers both static and small objects without introducing excessive false-positives. 
From here, we train the final detector from scratch, on the training pair $(I_i, \Ltwo)$ to convergence.

\subsection{Implementation details} 
\label{sec:method-implementation}
We follow the official code release by Dynamo-Depth\cite{sun2024dynamo} and train the system on Waymo Open \cite{waymo}.
During initial pseudo-label generation, we binarize the estimated motion mask via a threshold of $0.1$ and cluster the pseudo 3D points $P_i$ via DBSCAN \cite{dbscan} using a 10-by-10 local pixel neighborhood connectivity.

We adopt Mask R-CNN \cite{he2017mask} with a ResNet-50\cite{he2016deep} backbone as the detector architecture. We initialize the backbone via two strategies, namely MoCo v2\cite{MoCov2} on randomly sampled Waymo~\cite{waymo} patches and MoCo v2 on ImageNet \cite{imagenet}, denoted as \ours{}$^\ddag$ and \ours{}, respectively.
We use Adam optimizer~\cite{adam} with initial learning rate of 1e-4 and decay by $\frac12$ after 10 epochs.

During \mm{}, we train a detector for 3 epochs, with scale jittering from $0.5$ to $1.0$. Since early-stopping is applied, we adopt a lower confidence threshold of $0.5$ to compute the next round pseudo-labels $\Lone$.
During \ls{}, we train both the large and small detectors for 20 epochs, with fixed scaling at $1.0$ and $0.25$, respectively. As both are trained to convergence, we adopt a higher confidence threshold of $0.9$ and $0.8$, respectively, to compute the next round pseudo-labels $\Ltwo$ with aggregation.
For \textit{Final round}, we train the detector from scratch for 20 epochs, with scale jittering from $0.5$ to $1.0$. The self-training in \ours{} takes 27 hours on 1 NVIDIA A6000 GPU.

\section{Experiments}
\label{sec:exp}

\subsection{Experimental Setup}

\textbf{Datasets.} 
For evaluation, we focus our attention to self-driving datasets since uncurated, unlabeled video data is available and detecting mobile objects is of interest for autonomous vehicles.
We train both \ours{}$^\ddag$ and \ours{} on Waymo~\cite{waymo} and compare our performance with baselines on Waymo.
We then also evaluate generalization to nuScenes \cite{nuscenes}, KITTI \cite{kitti}, and COCO \cite{COCO}. 

We split the 798 sequences from Waymo train set into 762 for training and 36 for validation. After method development concludes, we evaluate on the held-out 1,881 test images (averaging 28.4 mobile instances per image) from Waymo val set \cite{waymo_panoptic}. 
Additionally, nuScenes, KITTI, and COCO are only used for evaluating generalization. We test on 3,249 front-camera images (average of 8.2 mobile instances per image) from the nuImage validation set for nuScene and 7,481 images (average of 6.9 mobile instances per image) in the 2D Detection training set for KITTI. For COCO, we evaluate on 870 images (average of 3.8 mobile instances per image) in COCO val 2017 that contain ground vehicles.

\textbf{Baselines.}
For the task of unsupervised mobile object detection, there is no directly comparable baselines to the best of our knowledge. Therefore, we consider methods for unsupervised object detection, namely CutLER \cite{cutler} and HASSOD \cite{hassod}, as the closest points of comparison. 

\textit{CutLER} \cite{cutler} uses normalized cuts on DINO features (trained on ImageNet) to generate pseudo labels that are used to train a
Cascade Mask R-CNN \cite{cai2018cascade} with ResNet-50 backbone.
We evaluate the official checkpoint.
Furthermore, we consider an additional baseline where $\Lone$ is directly predicted via CutLER, which effectively ablates the use of motion cues, as denoted by CutLER$^{L2S}$.

\textit{HASSOD} \cite{hassod} is a follow-up to \textit{CutLER}. It discovers objects on COCO \cite{COCO} via a hierarchical adaptive clustering of DINO features (trained from ImageNet).
The hierarchical clustering yields three ``levels'': objects, object parts, and object sub-parts. 
For consistency, we consider all three hierarchical levels for evaluation.
As before, we use its official released checkpoint.
Since \ours{}$^\ddag$ is trained on Waymo Open~\cite{waymo}, we also consider a version of HASSOD solely trained on Waymo, which we denote as HASSOD$^\dagger$.

CutLER\cite{cutler} and HASSOD\cite{hassod} are trained to detect all objects in an image regardless of their ability to move.
However, our evaluation and approach is focused on mobile objects.
We therefore also consider an ``oracle'' version of these baselines where we additionally remove any CutLER and HASSOD predictions that overlap by less than 0.1 in IoU with the ground truth instances.
These oracles, namely CutLER$^*$, CutLER$^{L2S*}$, HASSOD$^*$, and HASSOD$^{\dagger *}$, are grayed out in the tables to indicate additional ground-truth-based filtering.

We also consider a fully-supervised Mask R-CNN (Sup. Mask R-CNN) trained on COCO~\cite{COCO} for an oracle comparison, marked in gray.

\textbf{Metrics.}
We evaluate both Average Recall (AR$_{100}^{\text{Box}}$ and AR$_{100}^{\text{Mask}}$) and Average Precision (AP$^{\text{Box}}$ and AP$^{\text{Mask}}$).
Since the task is unsupervised and no semantic information is given during training, we follow prior arts \cite{cutler, hassod} and evaluate class-agnostic AR and AP by treating all predicted and ground truth instances as a single class of ``foreground'' or ``mobile'' objects.

\begin{table}[tb]
  \caption{Unsupervised Mobile Object Detection on Waymo Open Dataset \cite{waymo}. We report detection and segmentation metrics and note the training data (Train) and backbone initialization data (Init.), including ImageNet (IN), COCO, and Waymo Open (W). $^\dagger$ indicates manual replication with official released code, and $^*$ indicates the removal of  proposals with <0.1 IoU overlap with ground truth instances.}
    \label{tab:waymo}
  \centering
  \renewcommand{\arraystretch}{1}
    \renewcommand{\tabcolsep}{0.7mm}
    \resizebox{\linewidth}{!}{
  \begin{tabular}{@{}l c c |  ccccc | ccccc@{}}
    \toprule
    Method & Train & Init. &\multicolumn{5}{c | }{Box} &\multicolumn{5}{c}{Mask}\\
    \midrule
    \midrule
    &  &  & AR$^{0.5}$ & AR & AR$_\text{S}$ & AR$_\text{M}$ & AR$_\text{L}$ & AR$^{0.5}$ & AR & AR$_\text{S}$ & AR$_\text{M}$ & AR$_\text{L}$\\
    \midrule
    \multicolumn{2}{l}{\gray Sup. Mask R-CNN~\cite{he2017mask}} & & \gray 54.3  & \gray 31.9  & \gray 13.4  & \gray 53.0  & \gray 79.8  & \gray 48.9  & \gray 27.5  & \gray 9.2   & \gray 46.3  & \gray 78.9 \\   
    \midrule                                                                                                                                    
    CutLER\cite{cutler} &IN &IN        & 20.9  & 11.7  & 1.3   & 17.3  & 54.1  & 20.0  & 10.7  & 0.9   & 14.9  & 52.7 \\
    CutLER$^{L2S}$\cite{cutler} &IN+W &IN & 29.3  & 15.0  & 4.6   & 24.3  & 48.2  & 28.1  & 14.4  & 4.2   & 22.9  & 47.8 \\
    HASSOD\cite{hassod} &COCO &IN      & 21.9  & 12.7  & 1.8   & 17.6  & \textbf{59.5}  & 20.7  & 11.0  & 1.2   & 14.2  &\textbf{ 55.6} \\
    HASSOD$^\dagger$\cite{hassod} &W &IN       & 15.3  & 8.3   & 0.5   & 11.0  & 43.7  & 14.6  & 7.2   & 0.2   & 7.9   & 42.6 \\
    \ours{}$^\ddag$ (ours) &W &W             & \textbf{40.0}  & 17.5  & 8.4   & 25.7  & 46.4  & 35.4  & 14.6  & 6.8   & 21.2  & 40.4 \\
    \ours{} (ours) &W &IN           & 39.9  & \textbf{19.3}  & \textbf{8.6}   & \textbf{28.5}  & 54.1  & \textbf{35.8}  & \textbf{16.4}  & \textbf{7.2}   & \textbf{23.8}  & 47.3 \\
    \midrule
    \midrule
    &  &  & AP$_{50}$ & AP & AP$_\text{S}$ & AP$_\text{M}$ & AP$_\text{L}$ & AP$_{50}$ & AP & AP$_\text{S}$ & AP$_\text{M}$ & AP$_\text{L}$\\
    \midrule
    \multicolumn{2}{l}{\gray Sup. Mask R-CNN~\cite{he2017mask}} & & \gray 46.1  & \gray 25.7  & \gray 8.0   & \gray 42.2  & \gray 72.2  & \gray 41.9  & \gray 22.4  & \gray 5.3   & \gray 36.2  & \gray 73.0 \\   
    \midrule                                                                                                                                    
    CutLER\cite{cutler} &IN &IN        & 8.8   & 5.0   & 0.5   & 3.9   & 32.0  & 9.1   & 5.2   & 0.0   & 3.4   & 34.6 \\
    CutLER$^{L2S}$\cite{cutler} &IN+W &IN &  9.6   &  4.3   &  0.9   &  9.6   &  16.9  &  9.6   &  4.4   &  0.8   &  9.8   &  17.7 \\
    HASSOD\cite{hassod} &COCO &IN      & 5.0   & 3.1   & 0.1   & 2.5   & 28.4  & 5.0   & 2.8   & 0.0   & 2.0   & 27.7 \\
    HASSOD$^\dagger$\cite{hassod} &W &IN       & 3.7   & 2.2   & 0.0   & 1.1   & 17.2  & 3.9   & 2.0   & 0.0   & 0.9   & 18.3 \\
    \ours{}$^\ddag$ (ours) &W &W             & 26.1  & 10.9  & 4.2   & 16.2  & 32.0  & 25.0  & 9.5   & 3.7   & 14.1  & 28.7 \\
    \ours{} (ours) &W &IN           & \textbf{26.3}  & \textbf{12.6}  & \textbf{4.9}   & \textbf{17.9}  & \textbf{41.0}  & \textbf{25.1}  & \textbf{11.1}  & \textbf{4.5}   & \textbf{15.6}  & \textbf{36.3} \\
    \specialrule{0.01pt}{2pt}{2pt}
    \gray CutLER$^*$\cite{cutler} & \gray IN & \gray IN      & \gray 14.3  & \gray 7.4   & \gray 1.3   & \gray 10.7  & \gray 38.4  & \gray 14.3  & \gray 7.4   & \gray 0.9   & \gray 9.6   & \gray 41.0 \\
    \gray CutLER$^{L2S*}$\cite{cutler} & \gray IN+W & \gray IN    & \gray 23.2  & \gray 10.1  & \gray 3.3   & \gray 16.4  & \gray 32.6  & \gray 22.8  & \gray 10.2  & \gray 3.1   & \gray 16.3  & \gray 34.1 \\
    \gray HASSOD$^*$\cite{hassod} & \gray COCO & \gray IN    & \gray 15.3  & \gray 8.5   & \gray 1.4   & \gray 10.9  & \gray 42.6  & \gray 15.0  & \gray 7.7   & \gray 1.0   & \gray 8.9   & \gray 41.3 \\
    \gray HASSOD$^\dagger$$^*$\cite{hassod} & \gray W & \gray IN     & \gray 9.3   & \gray 4.7   & \gray 0.6   & \gray 6.5   & \gray 25.2  & \gray 9.6   & \gray 4.5   & \gray 0.1   & \gray 4.9   & \gray 27.0 \\
    \bottomrule
  \end{tabular}
  }
\end{table}

\begin{table}[tb]
  \caption{Zero-shot Unsupervised Mobile Object Detection on nuScenes \cite{nuscenes}. We report detection and segmentation metrics and note the training data (Train) and backbone initialization data (Init.), including ImageNet (IN), COCO, and Waymo Open (W). $^\dagger$ indicates manual replication with official released code, and $^*$ indicates the removal of  proposals with <0.1 IoU overlap with ground truth instances.}
    \label{tab:nuscenes}
  \centering
  \renewcommand{\arraystretch}{1}
    \renewcommand{\tabcolsep}{0.7mm}
    \resizebox{\linewidth}{!}{
  \begin{tabular}{@{}l c  c  |  ccccc  |  ccccc@{}}
    \toprule
    Method & Train & Init. &\multicolumn{5}{c | }{Box} &\multicolumn{5}{c}{Mask}\\
    \midrule
    \midrule
    &  &  & AR$^{0.5}$ & AR & AR$_\text{S}$ & AR$_\text{M}$ & AR$_\text{L}$ & AR$^{0.5}$ & AR & AR$_\text{S}$ & AR$_\text{M}$ & AR$_\text{L}$\\
    \midrule
    CutLER\cite{cutler}                   &IN &IN   & 28.9  & 16.0  & 3.8   & 23.4  & 56.1  & 27.8  & 14.3  & 3.0   & 20.4  & 53.0 \\
    HASSOD\cite{hassod}                   &COCO &IN & 30.9  & 17.0  & 5.0   & 24.0  & \textbf{56.8}  & 29.7  & 14.7  & 3.9   & 20.1  & \textbf{53.6} \\
    HASSOD$^\dagger$\cite{hassod}                 &W &IN    & 24.3  & 12.7  & 2.7   & 18.0  & 48.2  & 23.0  & 10.7  & 1.9   & 14.0  & 45.7 \\
    \ours{}$^\ddag$ (ours)        &W &W     & 42.1  & 17.3  & 8.7   & 24.2  & 39.8  & 36.4  & 13.9  & 8.2   & 18.1  & 29.7 \\
    \ours{} (ours)       &W &IN    & \textbf{48.9}  & \textbf{21.9}  & \textbf{12.0}  & \textbf{29.8}  & 48.2  & \textbf{42.3}  & \textbf{18.3}  & \textbf{10.7}  & \textbf{24.0}  & 39.1 \\
    \midrule
    \midrule
    &  &  & AP$_{50}$ & AP & AP$_\text{S}$ & AP$_\text{M}$ & AP$_\text{L}$ & AP$_{50}$ & AP & AP$_\text{S}$ & AP$_\text{M}$ & AP$_\text{L}$\\
    \midrule
    CutLER\cite{cutler}                   &IN &IN   & 6.0   & 3.7   & 0.3   & 3.1   & 23.7  & 5.9   & 3.5   & 0.1   & 2.8   & 22.7 \\
    HASSOD\cite{hassod}                   &COCO &IN & 3.9   & 2.2   & 0.1   & 2.1   & 20.5  & 3.8   & 2.0   & 0.1   & 1.8   & 19.5 \\
    HASSOD$^\dagger$\cite{hassod}                 &W &IN    & 3.6   & 2.2   & 0.0   & 1.7   & 15.2  & 3.6   & 1.8   & 0.0   & 0.9   & 14.6 \\
    \ours{}$^\ddag$ (ours)        &W &W     & 18.8  & 7.3   & 2.6   & 10.4  & 22.3  & 17.1  & 6.0   & 2.4   & 8.0   & 17.2 \\
    \ours{} (ours)       &W &IN    & \textbf{23.6}  & \textbf{10.7}  & \textbf{4.3}   & \textbf{14.9}  & \textbf{31.5}  & \textbf{21.8}  & \textbf{9.0}   & \textbf{3.8}   & \textbf{12.2}  & \textbf{25.6} \\
    \specialrule{0.01pt}{2pt}{2pt}
    \gray CutLER$^*$\cite{cutler}                 & \gray IN & \gray IN   & \gray  15.6  & \gray  8.2   & \gray  2.3   & \gray  12.6  & \gray  38.4  & \gray  15.2  & \gray  7.7   & \gray  1.7   & \gray  11.4  & \gray  37.0 \\
    \gray HASSOD$^*$\cite{hassod}                 & \gray COCO & \gray IN & \gray  18.6  & \gray  9.5   & \gray  2.3   & \gray  13.3  & \gray  38.2  & \gray  17.9  & \gray  8.6   & \gray  1.8   & \gray  11.5  & \gray  36.2 \\
    \gray HASSOD$^\dagger$$^*$\cite{hassod}               & \gray W & \gray IN    & \gray  13.0  & \gray  6.3   & \gray  1.2   & \gray  9.1   & \gray  27.9  & \gray  12.8  & \gray  5.6   & \gray  1.1   & \gray  7.5   & \gray  27.2 \\
    \bottomrule
  \end{tabular}
  }
\end{table}

\begin{table}[tb]
  \caption{Unsupervised Mobile Object Detection on KITTI \cite{kitti} and COCO \cite{COCO}. We report Average Recall (AR$_{100}^{\text{Box}}$) and Average Precision (AP$^{\text{Box}}$). Manual replication with official released code is indicated by $^\dagger$, and $^*$ indicates the removal of  proposal with less than 0.1 IoU overlap with all ground truth instances.}
    \label{tab:kitti_coco}
  \centering
  \renewcommand{\arraystretch}{1}
    \renewcommand{\tabcolsep}{0.7mm}
    \resizebox{\linewidth}{!}{
  \begin{tabular}{@{}l c  c  |  ccccc  |  ccccc@{}}
    \toprule
    Method & Train & Init. &\multicolumn{5}{c | }{KITTI \cite{kitti}} &\multicolumn{5}{c}{COCO \cite{COCO}}\\
    \midrule
    \midrule
    &  &  & AR$^{0.5}$ & AR & AR$_\text{S}$ & AR$_\text{M}$ & AR$_\text{L}$ & AR$^{0.5}$ & AR & AR$_\text{S}$ & AR$_\text{M}$ & AR$_\text{L}$\\
    \midrule
    CutLER\cite{cutler} &IN &IN                       & 50.9  & 24.4  & 11.2  & 23.0  & 42.9  & 49.0  & \textbf{27.9}  & 9.8   & 33.9  & \textbf{61.6} \\
    HASSOD\cite{hassod} &COCO &IN                     & 52.4  & 26.4  & 13.4  & 23.9  & \textbf{47.1}  & 51.2  & \textbf{27.9}  & 12.5  & \textbf{36.2}  & 51.7 \\
    HASSOD$^\dagger$\cite{hassod} &W &IN              & 49.9  & 23.7  & 15.0  & 22.1  & 37.4  & 48.8  & 26.1  & 12.5  & 31.2  & 50.7 \\
    \ours{}$^\ddag$ (ours) &W &W                            & 66.1  & 29.3  & 21.2  & 29.0  & 39.6  & 55.9  & 23.3  & 20.1  & 26.9  & 25.5 \\
    \ours{} (ours) &W &IN                          & \textbf{68.1}  & \textbf{32.6}  & \textbf{23.7}  & \textbf{32.0}  & 44.3  & \textbf{58.3}  & 26.6  & \textbf{22.3}  & 29.8  & 32.0 \\
    \midrule
    \midrule
    &  &  & AP$_{50}$ & AP & AP$_\text{S}$ & AP$_\text{M}$ & AP$_\text{L}$ & AP$_{50}$ & AP & AP$_\text{S}$ & AP$_\text{M}$ & AP$_\text{L}$\\
    \midrule
    CutLER\cite{cutler} &IN &IN                       & 18.6  & 8.6   & 0.4   & 5.6   & 24.0  & 9.8   & 5.6   & 0.4   & 2.4   & \textbf{21.6} \\
    HASSOD\cite{hassod} &COCO &IN                     & 14.3  & 7.2   & 0.3   & 5.1   & \textbf{29.5}  & 3.1   & 1.4   & 0.1   & 1.9   & 7.6  \\
    HASSOD$^\dagger$\cite{hassod} &W &IN              & 16.7  & 7.5   & 0.6   & 6.8   & 19.3  & 4.7   & 2.7   & 0.3   & 3.1   & 10.0 \\
    \ours{}$^\ddag$ (ours) &W &W                            & 38.5  & 15.8  & 9.8   & 15.3  & 26.2  & 14.1  & 5.8   & 5.1   & 7.6   & 5.9  \\
    \ours{} (ours) &W &IN                          & \textbf{38.9}  & \textbf{18.0}  & \textbf{11.7}  & \textbf{18.0}  & 28.4  & \textbf{14.2}  & \textbf{6.6}   & \textbf{5.6}   & \textbf{9.0}   & 6.6  \\
    \specialrule{0.01pt}{2pt}{2pt}
    \gray CutLER$^*$\cite{cutler} & \gray IN & \gray IN                   & \gray  28.4  & \gray  12.3  & \gray  3.5   & \gray  9.9   & \gray  28.8  & \gray  24.3  & \gray  12.8  & \gray  4.3   & \gray  12.1  & \gray  40.1 \\
    \gray HASSOD$^*$\cite{hassod} & \gray COCO & \gray IN                 & \gray  33.4  & \gray  15.4  & \gray  4.4   & \gray  11.5  & \gray  34.9  & \gray  21.3  & \gray  10.1  & \gray  5.4   & \gray  13.8  & \gray  18.9 \\
    \gray HASSOD$^\dagger$$^*$\cite{hassod} & \gray W & \gray IN          & \gray  30.8  & \gray  12.7  & \gray  5.7   & \gray  11.1  & \gray  23.3  & \gray  22.8  & \gray  10.6  & \gray  5.8   & \gray  12.7  & \gray  21.3 \\
    \bottomrule
  \end{tabular}
  }
\end{table}

\begin{figure}[hbt!]
  \centering
  \includegraphics[width=\textwidth]{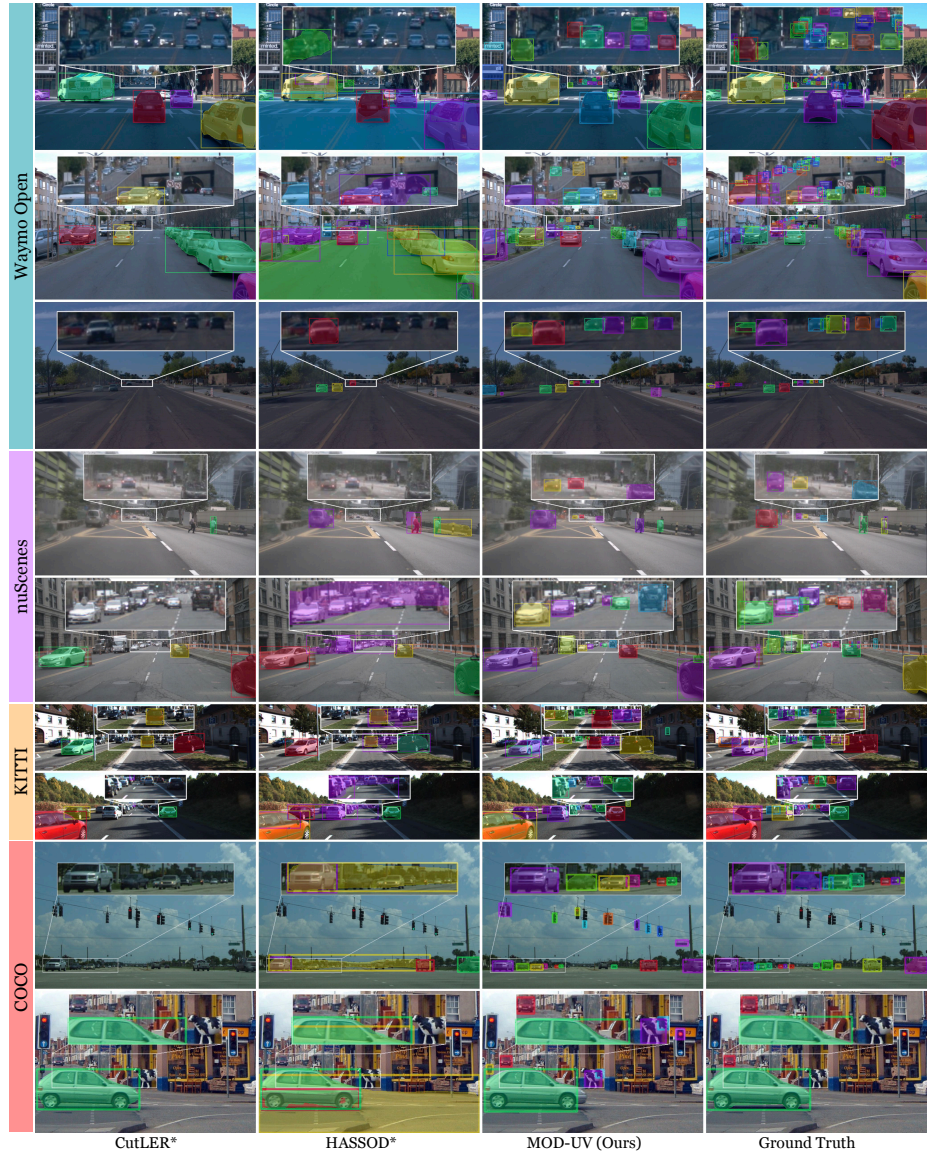}
  \caption{Qualitative Results on Waymo Open, nuScenes, KITTI, and COCO, where all proposals with over 0.5 confidence are visualized. For CutLER and HASSOD, we apply an additional filtering that removes any proposals with <0.1 IoU with ground truth mobile objects, as denoted by CutLER$^*$ and HASSOD$^*$, respectively.}
  \label{fig:qual}
\end{figure}
\begin{table}[tb]
  \caption{Ablation Study on the processing of pseudo-labels with \ours{}$^\ddag$. We report pseudo-label mask quality in terms of AR$^{\text{Mask}}$ on the training set of Waymo Open \cite{waymo}.}
    \label{tab:ablate_pseudo_mask}
  \centering
  \renewcommand{\arraystretch}{1}
    \renewcommand{\tabcolsep}{0.7mm}
    \resizebox{\linewidth}{!}{
  \begin{tabular}{@{}c c  |  cccc  |  cccc  |  cccc@{}}
    \toprule
    Pseudo & \# Epochs  &\multicolumn{4}{c | }{All} &\multicolumn{4}{c | }{Static} &\multicolumn{4}{c}{Moving}\\
    Masks & in \textit{M2M} & AR & AR$_\text{S}$ & AR$_\text{M}$ & AR$_\text{L}$ & AR & AR$_\text{S}$ & AR$_\text{M}$ & AR$_\text{L}$ & AR & AR$_\text{S}$ & AR$_\text{M}$ & AR$_\text{L}$\\
    \midrule
    Contour & \xmark     & 0.9 & 0.0 & 0.3 & 6.4 & 0.6 & 0.0 & 0.0 & 2.0 & 3.8 & 0.0 & 1.0 & 14.4 \\
    Depth & \xmark     & 1.5 & 0.0 & 0.5 & 10.0 & 0.7 & 0.0 & 0.0 & 2.4 & 6.3 & 0.0 & 1.5 & 23.8 \\
    \midrule
    Depth & 1               & 4.4 & 0.0 & 3.2 & 25.9 & 7.1 & 0.0 & 2.2 & 21.1 & 11.4 & 0.0 & 6.7 & 34.7 \\
    Depth & 3               & 5.3 & 0.0 & 4.6 & 29.0 & 9.1 & 0.0 & 3.5 & 25.8 & 12.4 & 0.0 & 8.7 & 35.1 \\
    Depth & 5               & 4.3 & 0.0 & 3.2 & 24.7 & 6.7 & 0.0 & 1.8 & 20.2 & 11.1 & 0.0 & 7.1 & 32.9 \\
    Depth & 20              & 4.1 & 0.0 & 2.3 & 25.6 & 6.6 & 0.0 & 1.3 & 20.8 & 10.6 & 0.0 & 5.2 & 34.4 \\
  \bottomrule
  \end{tabular}
  }
\end{table}

\begin{table}[tb]
  \caption{Ablation Study on the proposed self-training scheme involving \mm{}, \ls{}, and \textit{Final Round} with \ours{}$^\ddag$. We report Box AR$_{100}$ and Box AP for Unsupervised Mobile Object Detection on the Waymo Open~\cite{waymo}.}
    \label{tab:ablate_st}
  \centering
  \renewcommand{\arraystretch}{1}
    \renewcommand{\tabcolsep}{0.7mm}
    \resizebox{\linewidth}{!}{
  \begin{tabular}{@{}c c c c c |  ccccc  |  ccccc@{}}
    \toprule
     Stage 1  & & Stage 2 &  & Stage 3 & \multicolumn{5}{c | }{Box AR} & \multicolumn{5}{c}{Box AP}\\ 
    \textit{M2M}  & $\xrightarrow{}$ & \textit{L2S} & $\xrightarrow{}$ & \textit{Final} & AR$^{0.5}$ & AR & AR$_\text{S}$ & AR$_\text{M}$ & AR$_\text{L}$ &AP$_{50}$ & AP & AP$_\text{S}$ & AP$_\text{M}$ & AP$_\text{L}$\\
    \midrule
    \xmark && \xmark  && \cmark                           & 18.0  & 7.8   & 0.3   & 9.7   & 43.3  & 10.5  & 4.2   & 0.4   & 3.5   & 25.3 \\
    \xmark && $\dL$ only  && \xmark                      & 12.9  & 5.6   & 0.0   & 4.5   & 38.3  & 7.6   & 2.9   & 0.4   & 1.1   & 22.3 \\
    \xmark && $\dS$ only  && \xmark                      & 23.0  & 10.2  & 3.0   & 18.3  & 28.3  & 11.8  & 5.1   & 1.6   & 8.9   & 15.0 \\
    \xmark && $\dL+\dS$  && \cmark                          & 28.4  & 12.6  & 2.3   & 20.8  & 47.7  & 16.4  & 7.0   & 1.6   & 10.6  & 32.8 \\
    \midrule
    \cmark && \xmark  && \xmark                           & 23.2  & 10.1  & 1.0   & 16.0  & 45.1  & 13.7  & 5.6   & 0.5   & 7.8   & 28.1 \\
    \cmark && \xmark  && \cmark                           & 28.1  & 12.5  & 2.2   & 20.3  & 48.4  & 16.7  & 7.1   & 1.4   & 10.8  & 33.5 \\
    \cmark && $\dL$ only  && \xmark                           & 21.3  & 9.4   & 0.2   & 14.9  & 45.2  & 12.8  & 5.7   & 0.5   & 7.2   & 31.2 \\
    \cmark && $\dS$ only  && \xmark                           & 31.0  & 13.4  & 5.8   & 22.4  & 32.5  & 17.1  & 7.2   & 2.9   & 13.6  & 15.0 \\
    \cmark && $\dL+\dS$ && \cmark                               & 40.0  & 17.5  & 8.4   & 25.7  & 46.4  & 26.1  & 10.9  & 4.2   & 16.2  & 32.0 \\
  \bottomrule
  \end{tabular}
  }
\end{table}

\subsection{Unsupervised Mobile Object Detection and Segmentation}

\subsubsection{In-Domain Performance on Waymo.} 
\Cref{tab:waymo} compares \ours{}$^\ddag$ and \ours{} against CutLER, HASSOD, and HASSOD$^\dagger$ in unsupervised mobile object detection on Waymo. We also report performance for a MaskR-CNN trained on COCO as an oracle in the first row of \Cref{tab:waymo}.

\textbf{Recall.}
We report AR$^{0.5}$ (average recall with IoU$=0.5$), AR, \ARS{}, \ARM{}, and \ARL{} for both box and mask predictions. 
\ours{} significantly outperforms prior arts across all recall metrics except the recall for large objects where it is comparable.
The improvement is especially large for small objects (\textbf{4.7$\times$} higher AR$^{\text{Box}}_\text{S}$ than the nearest competitor).
Compared to a supervised Mask R-CNN trained on COCO~\cite{COCO}, \ours{} closes the gap in Box \ARS{} \textbf{from 11.6 to 4.8}.
Our gains are also much larger on the AR$^{0.5}$ metric (nearly \textbf{2$\times$} prior state-of-the-art). This suggests that we detect significantly more objects than prior work, but their localization can be improved.
Even so, we still show a $6$-point improvement on overall AR.
We also found HASSOD to underperform when trained solely on Waymo, which we suspect is due to the uncurated nature of self-driving scenes.

\textbf{Precision.}
We report AP at an overlap threshold of 0.5, as well as AP, \APS{}, \APM{} and \APL{}. 
Since CutLER and HASSOD are trained to detect all objects in an image regardless of their ability to move, we also compare to oracle versions of these techniques with ground-truth-based filtering.
On Waymo, \ours{} significantly outperforms prior arts across \textbf{all} precision metrics.
Even with ground-truth filtering (in gray), \ours{} still consistently outperforms baselines on all precision metrics (except for larger objects where it is comparable). 
Specifically, \ours{} outperforms the nearest competitor (with ground-truth filtering) by 4.1 on Box AP and 3.4 on Mask AP, and is \textbf{4.5$\times$} higher on AP$^{\text{Mask}}_\text{S}$.
Intriguingly, compared to a supervised Mask R-CNN trained on COCO~\cite{COCO}, \ours{} closes the gap in Mask \APS{} \textbf{from 4.3 to 0.8 points}.

\subsubsection{Generalization to Out-of-Domain Data.} 
We next take our detector trained on Waymo, and apply it \emph{out of the box} on nuScenes, KITTI, and COCO.

\textbf{Recall.} As shown in \Cref{tab:nuscenes} and \Cref{tab:kitti_coco}, on nuScenes and KITTI, \ours{} consistently outperforms prior arts across all AR metrics except for large objects, achieving a more than \textbf{1.5$\times$} improvement on AR$^{0.5}$ over the nearest competitor on nuScenes.
\ours{} also shows large gains on small objects, improving AR$^{\text{Box}}_\text{S}$ by \textbf{2.4$\times$} on nuScenes and over \textbf{1.7$\times$} on KITTI.

Finally, we evaluate on COCO, which is in-domain for HASSOD and a big domain shift for \ours{}. Notably, \ours{} maintains superiority on AR$_\text{S}$ and AR at IoU$=0.5$, while being comparable to HASSOD on AR and AR$_\text{M}$.

\textbf{Precision.} This improvement is also seen in AP.
On both nuScenes and KITTI, \ours{} consistently outperforms baseline on all AP metrics except being comparable for \APL with prior arts with ground-truth filtering. 
Specially, \ours{} improves upon HASSOD$^*$(with ground truth filtering) on Box AP by 1.2 on nuScenes and 2.6 on KITTI, with notable improvements on AP$^{\text{Box}}_\text{S}$ by over \textbf{1.8$\times$} on nuScenes and \textbf{2.6$\times$} on KITTI. 
Even on COCO, \ours{} outperforms prior arts on all metrics without ground-truth filtering except \APL{}.

\subsection{Qualitative Results}

In \Cref{fig:qual}, we show qualitative examples of \ours{} against CutLER and HASSOD after ground truth filtering, which we denote by CutLER$^*$ and HASSOD$^*$, respectively. In addition, we highlight the regions containing small objects with an additional zoom-in.

Without using any annotations, \ours{} detects mobile objects accurately, especially recovering many more small and faraway objects compared to prior arts. In contrast, due to the reliance on image features from static images, both CutLER and HASSOD tend to group multiple objects into a single proposal (seen in the second row in Waymo).

Notably, \ours{} reliably detects static and small mobile objects in the scene without excess amount of false positives. This improvement mostly originates from the proposed \mm{} and \ls{} (see ablation below).

Beyond accurate detection and segmentation on Waymo, \ours{} demonstrates impressive generalization when applied on nuScenes, KITTI, and COCO.

\subsection{Ablation Study}
\label{sec:exp-ablation}

We conduct ablation studies with \ours{}$^\ddag$ trained solely from Waymo~\cite{waymo} to understand the effects of each proposed component, including pseudo-label generation, static object discovery, small object discovery, and final round training.

\subsubsection{Motion Cues for Initial Pseudo-Labels.}
As shown in \Cref{tab:waymo}, there is a consistent AR improvement for small and medium objects in CutLER$^{L2S}$ over CutLER. 
This highlights the effectiveness of our \ls{} strategy in improving detector performance on small objects, which is a challenge for all unsupervised detectors/object discovery techniques because of the limited signal on small objects.
Despite of this gain, \ours{} still outperforms CutLER$^{L2S}$ because motion offers a stronger cue to separate small objects that appear close to each other in pixel space, as shown in \Cref{fig:qual}.

\subsubsection{Pseudo-Label Generation.}
In \Cref{tab:ablate_pseudo_mask}, we measure the quality of the pseudo-labels in terms of AR for All, the Static, and the Moving instances.

In the top of \Cref{tab:ablate_pseudo_mask}, compared to using 2D contours for generate pseudo-labels, clustering pseudo 3D points from monocular depth estimations improves Moving \ARL{} by 9.4 and Moving AR by 2.5.
This underlines the benefit in leveraging 3D information for partitioning moving instances.
Nevertheless, the Static AR is notably smaller, with Static \ARL{} only up to \textbf{10\%} of Moving \ARL{}.
Also, small and medium objects are almost entirely missed, with All \ARS{} at 0.0 and All \ARM{} at 0.5, compared to All \ARL{} at 10.0.
These observations further verify the bias pointed out in \Cref{sec:method-init_with_mask}, where static and small objects are mostly absent in the initial pseudo-labels generated from motion segmentation.

Furthermore, in the bottom half of \Cref{tab:ablate_pseudo_mask}, we demonstrate the effectiveness of self-training in the \mm{} stage in recovering static objects from the initial pseudo-labels with varying number of training epochs.
It is worth noting that regardless of convergence, self-training successfully improves Static AR, with improvements as high as 23.4 on Static \ARL{}.
Interestingly, as the initial pseudo-labels contain mostly moving objects, additional training beyond 3 epochs shows clear degradation in performance (reducing Static \ARL{} by 5.0 while retaining Moving \ARL{}), as the trained detector overfits to moving instances by exploiting contextual priors. 
In addition to improvements on All \ARL{}, the \mm{} stage is also able to slightly lift up All \ARM{} by the highest at 4.1. Nevertheless, with no improvements on \ARS{}, it is clear that the \mm{} stage alone cannot alleviate the negative bias from motion segmentation.

\subsubsection{Self-Training Pipeline.}
In \Cref{tab:ablate_st}, we evaluate every combination of the 3-stage self-training pipeline with \ours{}$^\ddag$ to evaluate the effectiveness of each.
Here, the \mm{} stage is again shown to be essential for recovering static objects from the initial pseudo-labels.
When \mm{} is ablated, performance decrease across all combinations, with notable reductions in AR by 4.9 and AP by 3.9 when solely ablated from \ours{}$^\ddag$.

Additionally, when \ls{} is ablated, \ARS{} reduces by nearly \textbf{4$\times$} and \APS{} by \textbf{3$\times$}, underlining its importance for small object detection.
Lastly, the final self-training round effectively learns the aggregated proposals from \ls{}, leading to an improved Box AR by 8.1 from $\dL$arge-object detector trained at original scale and by 4.1 from $\dS$mall-object detector trained at reduced scale.

\section{Conclusion}
\label{sec:conclusion}

We argue that motion is an important cue for unsupervised object detection, and propose the task of unsupervised mobile object detection. We propose a new training pipeline, \ours{}, that bootstraps from motion segmentation but removes its bias by discovering static and small objects. \ours{} achieves significant improvement over prior self-supervised detectors on multiple datasets.

\textbf{Limitations.} 
Our work makes an assumption that all mobile objects would often move in the given unlabeled video dataset. 
Although \ours{} can ideally learn and detect all mobile objects, in practice, the learning-based framework can only learn and detect things that frequently move in the videos. 
That said, for general applications where the autonomous agents can manipulate their surroundings, the agent can still learn to detect rarely moving objects by interacting with their environment, \eg poking at static objects within reach.

\textbf{Societal Impact.}
Being an unsupervised detection framework, our work does not include any negative social impacts beyond object detection itself.
\section*{Acknowledgement}

This research is based upon work supported in part by the National Science Foundation (IIS-2144117 and IIS-2107161).
Yihong Sun is supported by an NSF graduate research fellowship.

\bibliographystyle{splncs04}
\bibliography{main}

\newpage
\appendix

\newcommand{\ml}{\bm{M_L}} 
\newcommand{\ms}{\bm{M_S}} 

\section{Implementation Details}
\label{app:impl}
\subsection{Mask Aggregation}
\label{app:impl-mask}

When aggregating the predictions of ``large object'' detector, $\ml$, and ``small object'' detector, $\ms$, we found that $\ms$ contains mostly parts of large objects and small objects, while $\ml$ contains mostly large objects and groups of small objects. 
Intuitively, NMS is less-suited for this aggregation task due to the presence of object parts and groups.

Thus, we implement our aggregation as shown in Algo.~\ref{algo:agg}. Here, we first filter out smaller overlapping masks in $\ml$ (\eg ones that are covered by larger masks by more than \texttt{filtFrac} of 0.75) as the smaller objects should be found via $\ms$ instead. We also filter out larger overlapping masks in $\ms$ with the same \texttt{filtFrac} as the larger objects should be found via $\ml$ instead.

After directly matching the masks in $\ml$ and $\ms$ with a \texttt{matchThrd} of 0.5, if a subset of proposals in $\ms$ sufficiently covers a mask in $\ml$ (over a \texttt{coverFrac} of 0.5), then we consider the large proposal to likely be a group of instances and only keep the subset. Conversely, we consider the subset to likely be object parts and keep the large proposal. 

As shown in \Cref{tab:ablate_agg}, our aggregation approach improves upon NMS with a \texttt{matchThrd} of 0.5 by 1.8 in AR$^{0.5}$ and 1.5 in AP$_\text{50}$.

\begin{algorithm}
\caption{Masks Aggregation}
\begin{algorithmic}[1]
\Procedure{MaskAgg}{\texttt{ML}, \texttt{MS}, \texttt{matchThrd}, \texttt{filtFrac}, \texttt{coverFrac}}
    \State \textit{Description: Aggregate the large masks \texttt{ML} with the small masks \texttt{MS}} 
    \\
    \State \texttt{ML} $\xleftarrow{}$ \textsc{RemoveSmallerOverlappingMasks}(\texttt{ML}, \texttt{filtFrac}) 
    \State \texttt{MS} $\xleftarrow{}$ \textsc{RemoveLargerOverlappingMasks}(\texttt{MS}, \texttt{filtFrac}) 
    \State \texttt{MAgg} $\xleftarrow{}$ \texttt{\{\}} 
    
    \For{$\texttt{m} \in \texttt{ML}$}
        \State \texttt{mS} $\xleftarrow{}$ \textsc{OverlapSubset}(\texttt{MS},\texttt{m})
        \If{\texttt{mS} is empty}
            \State \textbf{continue}
        \ElsIf{\texttt{mS} is singulton $\&$ \textsc{IoU}(\texttt{mS}, \texttt{m}) > \texttt{matchThrd}}
            \State \texttt{MAgg} $\xleftarrow{}$ \texttt{MAgg} $\cup$ \textsc{HigherScoring}(\texttt{mS}, \texttt{m})
        \ElsIf{\textsc{Coverage}(\texttt{mS},\texttt{m}) > \texttt{coverFrac}}
            \State \texttt{MAgg} $\xleftarrow{}$ \texttt{MAgg} $\cup$ \texttt{mS}
        \Else
            \State \texttt{MAgg} $\xleftarrow{}$ \texttt{MAgg} $\cup$ \texttt{m}
        \EndIf
    \EndFor
    \\
    \State \texttt{MAgg} $\xleftarrow{}$ \texttt{MAgg} $\cup$ $\{ \texttt{m} : \texttt{m} \in \texttt{ML} \text{ \& } \textsc{Coverage}(\texttt{MS}, \texttt{m}) = 0\}$
    \State \texttt{MAgg} $\xleftarrow{}$ \texttt{MAgg} $\cup$ $\{ \texttt{m} : \texttt{m} \in \texttt{MS} \text{ \& } \textsc{Coverage}(\texttt{ML}, \texttt{m}) = 0\}$
    \\
    \State\algorithmicreturn~\texttt{MAgg}
\EndProcedure
\\
\Function{OverlapSubset}{\texttt{masks}, \texttt{targ\_mask}}
    \State \textit{Description: Find the subset of \texttt{masks} that overlap with \texttt{targ\_mask}} 
    \State\algorithmicreturn~ $\{ \texttt{m} : \texttt{m} \in \texttt{masks} \text{ \& } \textsc{Coverage}(\texttt{m}, \texttt{targ\_mask}) > 0\}$
\EndFunction
\\
\Function{Coverage}{\texttt{ref\_masks}, \texttt{targ\_mask}}
    \State \textit{Description: Return the fraction of \texttt{targ\_mask} that is covered by \texttt{ref\_masks}} 
    \State\algorithmicreturn~\text{IntersectArea}(\texttt{ref\_masks}, \texttt{targ\_mask}) / \text{Area}(\texttt{targ\_mask})
\EndFunction
\\
\Function{HigherScoring}{\texttt{m1}, \texttt{m2}}
    \State \textit{Description: Return the mask proposal with higher confidence score} 
    \State\algorithmicreturn~\texttt{m1} \textbf{if} \texttt{m1.score} > \texttt{m2.score} \textbf{else} \texttt{m2}
\EndFunction
\end{algorithmic}
\label{algo:agg}
\end{algorithm}

\section{Evaluation Datasets}
\label{app:eval-datasets}
\subsection{Waymo Open Dataset}

We evaluate performance on Waymo via all images in the val set of Waymo Open Dataset \cite{waymo}. We obtain the instance-level object masks via the panoptic annotations \cite{waymo_panoptic}, and treat the following object categories to be mobile:
\texttt{car}, \texttt{truck}, \texttt{bus}, \texttt{other\_vehicle}, \texttt{bicycle}, \texttt{motorcycle}, \texttt{trailer}, \texttt{pedestrian}, \texttt{bicyclist}, \texttt{motorcyclist}, \texttt{bird}, and \texttt{ground\_animal}. 

In total, there are 1,881 test images with 53,387 mobile instances labeled.

\subsection{nuScenes Dataset}

We evaluate performance on nuScenes via all FRONT camera images in the val set of nuImage \cite{nuscenes}. We consider all object categories under the super-categories (\texttt{animal},\texttt{human}, and \texttt{vehicle}) to be mobile. In total, there are 3,249 test images with 26,618 mobile instances labeled.

\subsection{KITTI Dataset}

We evaluate performance on KITTI via all images in the train set of KITTI 2D Detection Dataset \cite{kitti}. We consider all object categories labeled in the dataset to be mobile. In total, there are 7,481 test images with 51,865 mobile instances labeled.

\subsection{COCO Dataset}

We evaluate performance on COCO via all images that contain street vehicles in the val2017 set \cite{COCO}. We consider the following object categories to be mobile: \texttt{car}, \texttt{truck}, \texttt{bus}, \texttt{bicycle}, \texttt{motorcycle}.

In total, there are 870 test images with 3,319 mobile instances labeled. For reference, the original val2017 set contains 5,000 images with 36,781 labeled instances (regardless of mobility).

\begin{table}[tb]
  \caption{Unsupervised Mobile Object Detection on Waymo Open Dataset \cite{waymo} with complete metrics.}
    \label{tab:app-waymo}
  \centering
  \renewcommand{\arraystretch}{1.2}
    \renewcommand{\tabcolsep}{1.2mm}
    \resizebox{\linewidth}{!}{
  \begin{tabular}{@{}l c c | cccc|cccc | cccc|cccc@{}}
    \toprule
    Method & Train & Init. &\multicolumn{8}{c | }{Box} &\multicolumn{8}{c}{Mask}\\
    \midrule
     &  &  &\multicolumn{4}{c | }{\texttt{IoU=.50}} &\multicolumn{4}{c | }{\texttt{IoU=.50:.05:.95}} &\multicolumn{4}{c | }{\texttt{IoU=.50}} &\multicolumn{4}{c}{\texttt{IoU=.50:.05:.95}}\\
    \midrule
    \midrule
    &  &  & AR & AR$_\text{S}$ & AR$_\text{M}$ & AR$_\text{L}$ & AR & AR$_\text{S}$ & AR$_\text{M}$ & AR$_\text{L}$ & AR & AR$_\text{S}$ & AR$_\text{M}$ & AR$_\text{L}$ & AR & AR$_\text{S}$ & AR$_\text{M}$ & AR$_\text{L}$\\
    \midrule
    \multicolumn{2}{l}{ Sup. Mask R-CNN~\cite{he2017mask}} & & \gray 54.3  & \gray 33.2  & \gray 84.3  & \gray 95.1  & \gray 31.9  & \gray 13.4  & \gray 53.0  & \gray 79.8  & \gray 48.9  & \gray 25.1  & \gray 80.8  & \gray 96.9  & \gray 27.5  & \gray 9.2   & \gray 46.3  & \gray 78.9 \\
    \midrule
    CutLER\cite{cutler}  &  IN  &  IN  &  20.9  &  3.1  &  33.1  &  87.5  &  11.7  &  1.3  &  17.3  &  54.1  &  20.0  &  2.5  &  30.8  &  86.8  &  10.7  &  0.9  &  14.9  &  52.7\\
    CutLER$^{L2S}$\cite{cutler} &IN+W &IN &  29.3  &  10.7  &  47.9  &  85.1  &  15.0  &  4.6   &  24.3  &  48.2  &  28.1  &  9.6   &  45.3  &  84.7  &  14.4  &  4.2   &  22.9  &  47.8 \\
    HASSOD\cite{hassod}  &  COCO  &  IN  &  21.9  &  4.3  &  33.6  &  88.2  &  12.7  &  1.8  &  17.6  &  \textbf{59.5}  &  20.7  &  3.6  &  30.8  &  87.3  &  11.0  &  1.2  &  14.2  &  \textbf{55.6}\\
    HASSOD$^\dagger$\cite{hassod}  &  W  &  IN  &  15.3  &  1.1  &  22.9  &  73.2  &  8.3  &  0.5  &  11.0  &  43.7  &  14.6  &  0.8  &  20.2  &  74.6  &  7.2  &  0.2  &  7.9  &  42.6\\
    Ours  &  W  &  W  &  \textbf{40.0}  &  \textbf{23.0}  &  56.1  &  \textbf{92.5}  &  17.5  &  8.4  &  25.7  &  46.4  &  35.4  &  \textbf{19.0}  &  49.2  &  89.0  &  14.6  &  6.8  &  21.2  &  40.4\\
    Ours  &  W  &  IN  &  39.9  &  21.3  &  \textbf{59.9}  &  92.2  &  \textbf{19.3}  &  \textbf{8.6}  &  \textbf{28.5}  &  54.1  &  \textbf{35.8}  &  17.6  &  \textbf{53.2}  &  \textbf{90.0}  &  \textbf{16.4}  &  \textbf{7.2}  &  \textbf{23.8}  &  47.3\\
    \midrule
    \midrule
    &  &  & AP & AP$_\text{S}$ & AP$_\text{M}$ & AP$_\text{L}$ & AP & AP$_\text{S}$ & AP$_\text{M}$ & AP$_\text{L}$ & AP & AP$_\text{S}$ & AP$_\text{M}$ & AP$_\text{L}$ & AP & AP$_\text{S}$ & AP$_\text{M}$ & AP$_\text{L}$\\
    \midrule
    \multicolumn{2}{l}{ Sup. Mask R-CNN~\cite{he2017mask}} & & \gray 46.1  & \gray 22.6  & \gray 74.6  & \gray 91.5  & \gray 25.7  & \gray 8.0   & \gray 42.2  & \gray 72.2  & \gray 41.9  & \gray 15.9  & \gray 71.1  & \gray 94.7  & \gray 22.4  & \gray 5.3   & \gray 36.2  & \gray 73.0 \\
    \midrule
    CutLER\cite{cutler}  &  IN  &  IN  &  8.8  &  1.1  &  7.5  &  55.7  &  5.0  &  0.5  &  3.9  &  32.0  &  9.1  &  0.1  &  7.1  &  58.5  &  5.2  &  0.0  &  3.4  &  34.6\\
    CutLER$^{L2S}$\cite{cutler} &IN+W &IN &  9.6   &  2.0   &  20.9  &  35.6  &  4.3   &  0.9   &  9.6   &  16.9  &  9.6   &  1.8   &  20.8  &  36.3  &  4.4   &  0.8   &  9.8   &  17.7 \\
    HASSOD\cite{hassod}  &  COCO  &  IN  &  5.0  &  0.2  &  4.7  &  44.3  &  3.1  &  0.1  &  2.5  &  28.4  &  5.0  &  0.1  &  4.3  &  45.9  &  2.8  &  0.0  &  2.0  &  27.7\\
    HASSOD$^\dagger$\cite{hassod}  &  W  &  IN  &  3.7  &  0.0  &  2.3  &  29.6  &  2.2  &  0.0  &  1.1  &  17.2  &  3.9  &  0.0  &  2.0  &  33.8  &  2.0  &  0.0  &  0.9  &  18.3\\
    Ours  &  W  &  W  &  26.1  &  \textbf{12.3}  &  36.5  &  75.0  &  10.9  &  4.2  &  16.2  &  32.0  &  25.0  &  \textbf{11.0}  &  34.8  &  75.2  &  9.5  &  3.7  &  14.1  &  28.7\\
    Ours  &  W  &  IN  &  \textbf{26.3}  &  \textbf{12.3}  &  \textbf{37.9}  &  \textbf{76.5}  &  \textbf{12.6}  &  \textbf{4.9}  &  \textbf{17.9}  &  \textbf{41.0}  &  \textbf{25.1}  &  10.5  &  \textbf{35.9}  &  \textbf{77.5}  &  \textbf{11.1}  &  \textbf{4.5}  &  \textbf{15.6}  &  \textbf{36.3}\\
    \midrule
    CutLER$^*$\cite{cutler} &IN &IN                   & \gray 14.3  & \gray 2.8   & \gray 23.0  & \gray 69.5  & \gray 7.4   & \gray 1.3   & \gray 10.7  & \gray 38.4  & \gray 14.3  & \gray 2.1   & \gray 22.0  & \gray 72.2  & \gray 7.4   & \gray 0.9   & \gray 9.6   & \gray 41.0 \\
    CutLER$^{L2S*}$\cite{cutler} &  IN+W &  IN    & \gray 23.2  & \gray 8.5   & \gray 37.9  & \gray 68.4  & \gray 10.1  & \gray 3.3   & \gray 16.4  & \gray 32.6  & \gray 22.8  & \gray 7.7   & \gray 37.0  & \gray 69.7  & \gray 10.2  & \gray 3.1   & \gray 16.3  & \gray 34.1 \\
    HASSOD$^*$\cite{hassod} &COCO &IN                 & \gray 15.3  & \gray 3.0   & \gray 23.3  & \gray 68.6  & \gray 8.5   & \gray 1.4   & \gray 10.9  & \gray 42.6  & \gray 15.0  & \gray 2.3   & \gray 21.3  & \gray 70.6  & \gray 7.7   & \gray 1.0   & \gray 8.9   & \gray 41.3 \\
    HASSOD$^\dagger$$^*$\cite{hassod} &W &IN          & \gray 9.3   & \gray 1.3   & \gray 15.2  & \gray 46.4  & \gray 4.7   & \gray 0.6   & \gray 6.5   & \gray 25.2  & \gray 9.6   & \gray 0.3   & \gray 13.4  & \gray 52.7  & \gray 4.5   & \gray 0.1   & \gray 4.9   & \gray 27.0 \\
    \bottomrule
  \end{tabular}
  }
\end{table}

\begin{table}[tb]
  \caption{Zero-shot Unsupervised Mobile Object Detection on nuScenes \cite{nuscenes} with complete metrics.}
    \label{tab:app-nuscenes}
  \centering
  \renewcommand{\arraystretch}{1.2}
    \renewcommand{\tabcolsep}{1.2mm}
    \resizebox{\linewidth}{!}{
  \begin{tabular}{@{}l c c | cccc|cccc | cccc|cccc@{}}
    \toprule
    Method & Train & Init. &\multicolumn{8}{c | }{Box} &\multicolumn{8}{c}{Mask}\\
    \midrule
     &  &  &\multicolumn{4}{c | }{\texttt{IoU=.50}} &\multicolumn{4}{c | }{\texttt{IoU=.50:.05:.95}} &\multicolumn{4}{c | }{\texttt{IoU=.50}} &\multicolumn{4}{c}{\texttt{IoU=.50:.05:.95}}\\
    \midrule
    \midrule
    &  &  & AR & AR$_\text{S}$ & AR$_\text{M}$ & AR$_\text{L}$ & AR & AR$_\text{S}$ & AR$_\text{M}$ & AR$_\text{L}$ & AR & AR$_\text{S}$ & AR$_\text{M}$ & AR$_\text{L}$ & AR & AR$_\text{S}$ & AR$_\text{M}$ & AR$_\text{L}$\\
    \midrule
    CutLER\cite{cutler}  &  IN  &  IN  &  28.9  &  8.6  &  43.9  &  88.2  &  16.0  &  3.8  &  23.4  &  56.1  &  27.8  &  7.5  &  41.7  &  87.6  &  14.3  &  3.0  &  20.4  &  53.0\\
    HASSOD\cite{hassod}  &  COCO  &  IN  &  30.9  &  11.0  &  45.3  &  \textbf{90.2}  &  17.0  &  5.0  &  24.0  &  \textbf{56.8}  &  29.7  &  10.2  &  42.9  &  \textbf{88.2}  &  14.7  &  3.9  &  20.1  &  \textbf{53.6}\\
    HASSOD$^\dagger$\cite{hassod}  &  W  &  IN  &  24.3  &  6.2  &  36.3  &  81.6  &  12.7  &  2.7  &  18.0  &  48.2  &  23.0  &  5.2  &  33.8  &  80.5  &  10.7  &  1.9  &  14.0  &  45.7\\
    Ours  &  W  &  W  &  42.1  &  24.3  &  57.8  &  85.5  &  17.3  &  8.7  &  24.2  &  39.8  &  36.4  &  23.5  &  46.4  &  70.7  &  13.9  &  8.2  &  18.1  &  29.7\\
    Ours  &  W  &  IN  &  \textbf{48.9}  &  \textbf{31.8}  &  \textbf{65.0}  &  87.8  &  \textbf{21.9}  &  \textbf{12.0}  &  \textbf{29.8}  &  48.2  &  \textbf{42.3}  &  \textbf{27.4}  &  \textbf{54.8}  &  78.8  &  \textbf{18.3}  &  \textbf{10.7}  &  \textbf{24.0}  &  39.1\\
    \midrule
    \midrule
    &  &  & AP & AP$_\text{S}$ & AP$_\text{M}$ & AP$_\text{L}$ & AP & AP$_\text{S}$ & AP$_\text{M}$ & AP$_\text{L}$ & AP & AP$_\text{S}$ & AP$_\text{M}$ & AP$_\text{L}$ & AP & AP$_\text{S}$ & AP$_\text{M}$ & AP$_\text{L}$\\
    \midrule
    CutLER\cite{cutler}  &  IN  &  IN  &  6.0  &  0.6  &  6.6  &  37.4  &  3.7  &  0.3  &  3.1  &  23.7  &  5.9  &  0.2  &  6.1  &  37.3  &  3.5  &  0.1  &  2.8  &  22.7\\
    HASSOD\cite{hassod}  &  COCO  &  IN  &  3.9  &  0.3  &  4.4  &  33.6  &  2.2  &  0.1  &  2.1  &  20.5  &  3.8  &  0.2  &  4.1  &  33.1  &  2.0  &  0.1  &  1.8  &  19.5\\
    HASSOD$^\dagger$\cite{hassod}  &  W  &  IN  &  3.6  &  0.1  &  3.5  &  26.2  &  2.2  &  0.0  &  1.7  &  15.2  &  3.6  &  0.1  &  2.5  &  27.0  &  1.8  &  0.0  &  0.9  &  14.6\\
    Ours  &  W  &  W  &  18.8  &  7.4  &  27.3  &  53.6  &  7.3  &  2.6  &  10.4  &  22.3  &  17.1  &  7.3  &  23.7  &  45.4  &  6.0  &  2.4  &  8.0  &  17.2\\
    Ours  &  W  &  IN  &  \textbf{23.6}  &  \textbf{11.5}  &  \textbf{32.6}  &  \textbf{59.3}  &  \textbf{10.7}  &  \textbf{4.3}  &  \textbf{14.9}  &  \textbf{31.5}  &  \textbf{21.8}  &  \textbf{10.1}  &  \textbf{29.8}  &  \textbf{55.2}  &  \textbf{9.0}  &  \textbf{3.8}  &  \textbf{12.2}  &  \textbf{25.6}\\
    \midrule
    CutLER$^*$\cite{cutler} &IN &IN                   & \gray 15.6  & \gray 5.6   & \gray 26.8  & \gray 65.7  & \gray 8.2   & \gray 2.3   & \gray 12.6  & \gray 38.4  & \gray 15.2  & \gray 4.7   & \gray 25.2  & \gray 65.4  & \gray 7.7   & \gray 1.7   & \gray 11.4  & \gray 37.0 \\
    HASSOD$^*$\cite{hassod} &COCO &IN                 & \gray 18.6  & \gray 6.0   & \gray 28.8  & \gray 66.5  & \gray 9.5   & \gray 2.3   & \gray 13.3  & \gray 38.2  & \gray 17.9  & \gray 5.5   & \gray 26.8  & \gray 64.8  & \gray 8.6   & \gray 1.8   & \gray 11.5  & \gray 36.2 \\
    HASSOD$^\dagger$$^*$\cite{hassod} &W &IN          & \gray 13.0  & \gray 3.3   & \gray 21.3  & \gray 52.7  & \gray 6.3   & \gray 1.2   & \gray 9.1   & \gray 27.9  & \gray 12.8  & \gray 3.0   & \gray 19.8  & \gray 54.2  & \gray 5.6   & \gray 1.1   & \gray 7.5   & \gray 27.2 \\
    \bottomrule
  \end{tabular}
  }
\end{table}

\begin{table}[tb]
  \caption{Ablation Study on the mask aggregation step. We report Average Recall (AR$_{100}^{\text{Box}}$) and Average Precision (AP$^{\text{Box}}$) for Class-Agnostic Mobile Object Detection and Instance Segmentation on the Waymo Open Dataset \cite{waymo}.}
    \label{tab:ablate_agg}
  \centering
  \begin{tabular}{@{}c \hsp{5} | \hsp{2} ccccc \hsp{2} | \hsp{2} ccccc@{}}
    \toprule
    Mask Agg   &\multicolumn{5}{c\hsp{2} | \hsp{2}}{Box AR} &\multicolumn{5}{c}{Box AP}\\
    Strategies  & AR$^{0.5}$ & AR & AR$_\text{S}$ & AR$_\text{M}$ & AR$_\text{L}$ &AP$_{50}$ & AP & AP$_\text{S}$ & AP$_\text{M}$ & AP$_\text{L}$\\
    \midrule
    NMS                                            & 38.2  & 17.0  & 7.7   & 24.6  & 48.2  & 24.6  & 10.5  & 4.2   & 14.7  & 35.2 \\
    0.5/0.5                                        & 39.0  & 17.2  & 7.8   & 25.9  & 45.7  & 24.3  & 10.0  & 4.2   & 16.2  & 28.9 \\
    \midrule
    Ours                                        & 40.0  & 17.5  & 8.4   & 25.7  & 46.4  & 26.1  & 10.9  & 4.2   & 16.2  & 32.0 \\
  \bottomrule
  \end{tabular}
\end{table}

\begin{table}[tb]
  \caption{Unsupervised Mobile Object Detection on KITTI \cite{kitti} and COCO \cite{COCO} with complete metrics.}
    \label{tab:app-kitti_coco}
  \renewcommand{\arraystretch}{1.2}
    \renewcommand{\tabcolsep}{1.2mm}
    \resizebox{\linewidth}{!}{
  \centering
  \begin{tabular}{@{}l c c  |  cccc|cccc  |  cccc|cccc@{}}
    \toprule
    Method & Train & Init. &\multicolumn{8}{c}{KITTI \cite{kitti}} &\multicolumn{8}{c}{COCO \cite{COCO}}\\
    \midrule
     &  &  &\multicolumn{4}{c | }{\texttt{IoU=.50}} &\multicolumn{4}{c | }{\texttt{IoU=.50:.05:.95}} &\multicolumn{4}{c | }{\texttt{IoU=.50}} &\multicolumn{4}{c}{\texttt{IoU=.50:.05:.95}}\\
    \midrule
    \midrule
    &  &  & AR & AR$_\text{S}$ & AR$_\text{M}$ & AR$_\text{L}$ & AR & AR$_\text{S}$ & AR$_\text{M}$ & AR$_\text{L}$ & AR & AR$_\text{S}$ & AR$_\text{M}$ & AR$_\text{L}$ & AR & AR$_\text{S}$ & AR$_\text{M}$ & AR$_\text{L}$\\
    \midrule
    CutLER\cite{cutler}  &  IN  &  IN  &  50.9  &  25.0  &  50.4  &  83.3  &  24.4  &  11.2  &  23.0  &  42.9  &  49.0  &  19.3  &  64.9  &  \textbf{94.6}  &  \textbf{27.9}  &  9.8  &  33.9  &  \textbf{61.6}\\
    HASSOD\cite{hassod}  &  COCO  &  IN  &  52.4  &  29.6  &  50.5  &  83.9  &  26.4  &  13.4  &  23.9  &  \textbf{47.1}  &  51.2  &  24.1  &  \textbf{66.6}  &  91.7  &  \textbf{27.9}  &  12.5  &  \textbf{36.2}  &  51.7\\
    HASSOD$^\dagger$\cite{hassod}  &  W  &  IN  &  49.9  &  33.2  &  48.5  &  73.3  &  23.7  &  15.0  &  22.1  &  37.4  &  48.8  &  24.5  &  59.4  &  90.5  &  26.1  &  12.5  &  31.2  &  50.7\\
    Ours  &  W  &  W  &  66.1  &  49.5  &  67.6  &  83.4  &  29.3  &  21.2  &  29.0  &  39.6  &  55.9  &  48.7  &  61.3  &  64.8  &  23.3  &  20.1  &  26.9  &  25.5\\
    Ours  &  W  &  IN  &  \textbf{68.1}  &  \textbf{54.1}  &  \textbf{68.3}  &  \textbf{84.7}  &  \textbf{32.6}  &  \textbf{23.7}  &  \textbf{32.0}  &  44.3  &  \textbf{58.3}  &  \textbf{49.6}  &  62.9  &  72.1  &  26.6  &  \textbf{22.3}  &  29.8  &  32.0\\
    \midrule
    \midrule
    &  &  & AP & AP$_\text{S}$ & AP$_\text{M}$ & AP$_\text{L}$ & AP & AP$_\text{S}$ & AP$_\text{M}$ & AP$_\text{L}$ & AP & AP$_\text{S}$ & AP$_\text{M}$ & AP$_\text{L}$ & AP & AP$_\text{S}$ & AP$_\text{M}$ & AP$_\text{L}$\\
    \midrule
    CutLER\cite{cutler}  &  IN  &  IN  &  18.6  &  0.9  &  14.5  &  49.7  &  8.6  &  0.4  &  5.6  &  24.0  &  9.8  &  0.8  &  5.5  &  \textbf{37.2}  &  5.6  &  0.4  &  2.4  &  \textbf{21.6}\\
    HASSOD\cite{hassod}  &  COCO  &  IN  &  14.3  &  0.7  &  11.4  &  55.7  &  7.2  &  0.3  &  5.1  &  \textbf{29.5}  &  3.1  &  0.3  &  4.0  &  16.8  &  1.4  &  0.1  &  1.9  &  7.6\\
    HASSOD$^\dagger$\cite{hassod}  &  W  &  IN  &  16.7  &  1.4  &  16.2  &  42.5  &  7.5  &  0.6  &  6.8  &  19.3  &  4.7  &  0.6  &  5.4  &  21.0  &  2.7  &  0.3  &  3.1  &  10.0\\
    Ours  &  W  &  W  &  38.5  &  24.1  &  38.9  &  \textbf{60.0}  &  15.8  &  9.8  &  15.3  &  26.2  &  14.1  &  \textbf{12.0}  &  18.4  &  16.3  &  5.8  &  5.1  &  7.6  &  5.9\\
    Ours  &  W  &  IN  &  \textbf{38.9}  &  \textbf{27.0}  &  \textbf{39.2}  &  55.7  &  \textbf{18.0}  &  \textbf{11.7}  &  \textbf{18.0}  &  28.4  &  \textbf{14.2}  &  11.7  &  \textbf{19.1}  &  16.2  &  \textbf{6.6}  &  \textbf{5.6}  &  \textbf{9.0}  &  6.6\\
    \midrule
    CutLER$^*$\cite{cutler} &IN &IN                   & \gray 28.4  & \gray 9.2   & \gray 26.1  & \gray 61.7  & \gray 12.3  & \gray 3.5   & \gray 9.9   & \gray 28.8  & \gray 24.3  & \gray 9.9   & \gray 27.6  & \gray 70.3  & \gray 12.8  & \gray 4.3   & \gray 12.1  & \gray 40.1 \\
    HASSOD$^*$\cite{hassod} &COCO &IN                 & \gray 33.4  & \gray 12.1  & \gray 27.9  & \gray 67.7  & \gray 15.4  & \gray 4.4   & \gray 11.5  & \gray 34.9  & \gray 21.3  & \gray 12.3  & \gray 28.6  & \gray 40.6  & \gray 10.1  & \gray 5.4   & \gray 13.8  & \gray 18.9 \\
    HASSOD$^\dagger$$^*$\cite{hassod} &W &IN          & \gray 30.8  & \gray 15.3  & \gray 28.7  & \gray 53.1  & \gray 12.7  & \gray 5.7   & \gray 11.1  & \gray 23.3  & \gray 22.8  & \gray 13.5  & \gray 27.2  & \gray 46.3  & \gray 10.6  & \gray 5.8   & \gray 12.7  & \gray 21.3 \\
    \bottomrule
  \end{tabular}
  }
\end{table}

\section{Additional Quantitative Results}
\label{app:quant}
\subsection{Complete Metrics}
For additional performance metrics on Waymo Open \cite{waymo}, please refer to \Cref{tab:app-waymo} that corresponds to \Cref{tab:waymo}.
Also, nuScenes \cite{nuscenes} performances can be found in \Cref{tab:app-nuscenes} (corresponding to \Cref{tab:nuscenes}). Finally, KITTI \cite{kitti} and COCO \cite{COCO} results are found in \Cref{tab:app-kitti_coco} (corresponding to \Cref{tab:kitti_coco}).

Interestingly, we found that motion segmentation trained from a reconstruction objective \cite{sun2024dynamo} contains specific biases, such as the inclusion of object shadows and missing object parts due to smooth object regions. This results in localization errors and reduced performances for large objects for IoU>0.75. As shown in the \Cref{tab:app-waymo}, \ours{} outperforms all prior art even for large objects for IoU=0.5 on Waymo Open.

\subsection{Mask Aggregation}
In \Cref{tab:ablate_agg}, we evaluate the effects of different mask aggregation techniques on the performance of the final detector. 
Notably, different strategies have minimal effects on Box AR and Box AP, which suggests the robustness of the mask aggregation step.
Additionally, we do observe small performance gain for medium objects against the Non-Maximum Suppression (NMS) technique and small performance gain for large objects against the use of lowered confidence thresholds, ``0.5/0.5'' instead of ``0.9/0.8''.

\subsection{Backbone Pre-training.}
In \Cref{tab:ablate_backbone}, we evaluate the effects of different backbone initializations on our proposed learning scheme.
Since \ours{} depends on multiple rounds of self-training, training detectors with a backbone initialized from scratch ($\emptyset$) reduces performance significantly.
Interestingly, the use of any particular pre-training technique is less influential, as \ours{} demonstrate similar performances.

\begin{table}[tb]
  \caption{Ablation Study on the use of different datasets for initializing the ResNet-50 backbone. We report Average Recall (AR$_{100}^{\text{Box}}$) and Average Precision (AP$^{\text{Box}}$) for Class-Agnostic Mobile Object Detection and Instance Segmentation on the Waymo Open Dataset \cite{waymo}. Here, $\emptyset$, W, IN, and IN(L) denotes the use of encoder backbone from scratch, MoCov2 on Waymo patches, MoCov2 on ImageNet, and fully-supervised ImageNet classifiers, respectively.}
    \label{tab:ablate_backbone}
  \centering
  \begin{tabular}{@{}c \hsp{5} c \hsp{5} | \hsp{2} ccccc \hsp{2} | \hsp{2} ccccc@{}}
    \toprule
    Training & Init. Enc  &\multicolumn{5}{c\hsp{2} | \hsp{2}}{Box AR} &\multicolumn{5}{c}{Box AP}\\
    Data & Data & AR$^{0.5}$ & AR & AR$_\text{S}$ & AR$_\text{M}$ & AR$_\text{L}$ &AP$_{50}$ & AP & AP$_\text{S}$ & AP$_\text{M}$ & AP$_\text{L}$\\
    \midrule
    W &$\emptyset$                                        & 29.9  & 13.4  & 5.9   & 18.5  & 40.7  & 16.0  & 7.1   & 2.6   & 10.5  & 25.9 \\
    W &W                                        & 40.0  & 17.5  & 8.4   & 25.7  & 46.4  & 26.1  & 10.9  & 4.2   & 16.2  & 32.0 \\
    W &IN                                       & 39.9  & 19.3  & 8.6   & 28.5  & 54.1  & 26.3  & 12.6  & 4.9   & 17.9  & 41.0 \\
    W &IN(L)                                    & 39.9  & 18.7  & 9.1   & 27.5  & 48.6  & 26.3  & 11.8  & 5.3   & 17.5  & 33.6 \\
  \bottomrule
  \end{tabular}
\end{table}

\begin{table}[H]
  \caption{Repeated Experiments for Reproducibility. We report Average Recall (AR$_{100}^{\text{Box}}$) and Average Precision (AP$^{\text{Box}}$) for Class-Agnostic Mobile Object Detection and Instance Segmentation on the Waymo Open Dataset \cite{waymo}.}
    \label{tab:ablate_repeat}
  \centering
  \begin{tabular}{@{}c \hsp{5} | \hsp{2} ccccc \hsp{2} | \hsp{2} ccccc@{}}
    \toprule
    \multirow{2}{*}{Run}  &\multicolumn{5}{c\hsp{2} | \hsp{2}}{Box AR} &\multicolumn{5}{c}{Box AP}\\
     & AR$^{0.5}$ & AR & AR$_\text{S}$ & AR$_\text{M}$ & AR$_\text{L}$ &AP$_{50}$ & AP & AP$_\text{S}$ & AP$_\text{M}$ & AP$_\text{L}$\\
    \midrule
    Standard Deviation                                            & 1.3	& 0.8	& 0.8	& 1.1	& 3.5	& 0.8	& 0.9	& 0.3	& 1.3	& 4\\
  \bottomrule
  \end{tabular}
\end{table}

\subsection{Reproducibility and Hyperparameter Selection}
To ensure reproducibility, we repeat the experiment for \ours{}$^\ddag$ 3 times with randomly generated seeds and obtained a 95\% confidence interval of 16.9 $\pm$ 1.3 and 10.3 $\pm$ 1.5 for Box AR and Box AP, respectively on Waymo Open.
Please refer to \Cref{tab:ablate_repeat} for standard deviations for each metric.

The number of epochs for \mm{}, scale jittering rates, and confidence thresholds for self-training were found on a small Waymo validation set during the development of our paper, while the test set is held-out entirely. The rest of the hyperparameters, including number of epochs, learning rates, and decay, were set arbitrarily and not tuned since training converged.

\end{document}